\documentclass[preprint]{elsarticle}

\makeatletter
\def\ps@pprintTitle{%
	\let\@oddhead\@empty
	\let\@evenhead\@empty
	\let\@oddfoot\@empty
	\let\@evenfoot\@oddfoot
}
\makeatother
\pdfoutput=1	

\usepackage[utf8]{inputenc}

\usepackage{rotating} 
\usepackage{booktabs} 
\usepackage{wrapfig} 
\usepackage{caption} 
\usepackage{subcaption} 
\usepackage{makecell} 
\usepackage{amsmath}
\usepackage{amsfonts}
\usepackage{amssymb} 
\usepackage{bm} 
\usepackage{multicol}
\usepackage{xcolor}

\usepackage{amsthm} 

\theoremstyle{definition} 

\theoremstyle{remark}

\begin{document}

\begin{frontmatter}

\title{Identification and Visualization of the Underlying Independent Causes of the Diagnostic of Diabetic Retinopathy made by a Deep Learning Classifier}

\author[label1]{Jordi de la Torre\corref{cor1}}
\address[label1]{Departament d'Enginyeria Inform\`atica i Matem\`atiques.\\Escola T\`ecnica Superior d'Enginyeria.\\Universitat Rovira i Virgili\\Avinguda Paisos Catalans, 26. E-43007\\
	Tarragona, Spain}
\ead{jordi.delatorre@gmail.com}
\author[label1]{Aida Valls}
\ead{aida.valls@urv.cat}
\author[label1]{Domenec Puig}
\ead{domenec.puig@urv.cat}
\author[label2]{Pedro Romero-Aroca}
\ead{romeropere@gmail.com}

\cortext[cor1]{Corresponding author}

\address[label2]{Ophthalmic Service. University Hospital Sant Joan de Reus\\Institut d’Investigaci\'o Sanit\`aria Pere Virgili (IISPV)\\ Universitat Rovira i Virgili\\Reus (Tarragona)\\Avinguda de la Universitat, 1. E-43204\\Reus, Spain}

\date{May 31, 2018}

\begin{abstract}
Interpretability is a key factor in the design of automatic classifiers for medical diagnosis. Deep learning models have been proven to be a very effective classification algorithm when trained in a supervised way with enough data. The main concern is the difficulty of inferring rationale interpretations from them. Different attempts have been done in last years in order to convert deep learning classifiers from high confidence statistical black box machines into self-explanatory models. In this paper we go forward into the generation of explanations by identifying the independent causes that use a deep learning model for classifying an image into a certain class. We use a combination of Independent Component Analysis with a Score Visualization technique. In this paper we study the medical problem of classifying an eye fundus image into 5 levels of Diabetic Retinopathy. We conclude that only 3 independent components are enough for the differentiation and correct classification between the 5 disease standard classes. We propose a method for visualizing them and detecting lesions from the generated visual maps.
\end{abstract}


\end{frontmatter}


\section{Introduction}

Diabetic Retinopathy (DR) is a leading disabling chronic disease  and  one of the main causes of blindness and visual impairment in developed countries for diabetic patients. Studies reported that 90\% of the cases can be prevented through early detection and treatment. Eye screening through retinal images is used by physicians to detect the lesions related with this disease. Due to the increasing number of diabetic patients, the amount of images to be manually analyzed is becoming unaffordable. Moreover, training new personnel for this type of image-based diagnosis is long, because it requires to acquire expertise by daily practice \citep{pedro2010prevalence}, \citep{romero2011managing}.

Deep Learning (DL) \citep{nature-deep-learning}, \citep{Schmidhuber-nn}, is a subfield of Machine Learning that allow the automatic model construction of very effective image classifiers using a parametric model. These models are able to identify and extract the statistical regularities present in data that are important for optimizing a defined loss function, with the final objective of mapping a high-multidimensional input into a smaller multi-dimensional output (f: $\mathbb{R}^{n} \mapsto \mathbb{R}^{m}, n \gg m$). This mapping allows the classification of multi-dimensional objects into a small number of categories. The model is composed by many neurons that are organized in layers in a hierarchical way. Every neuron receives the input from a predefined set of neurons. Every connection has a parameter that corresponds to the weight of the connection. The function of every neuron is to make a transformation of the received inputs into a calculated output value. For every incoming connection, the weight is multiplied by the input value received by the neuron. The aggregation of all inputs passes through an activation function that calculates the neuron output. Parameters are usually optimized using a stochastic gradient descent algorithm that minimizes a predefined loss function. These hierarchical models are able to learn multiple levels of representation that correspond to different levels of abstraction, which enables the representation of complex concepts in a compressed way \citep{Bengio:2013:RLR:2498740.2498889}, \citep{bengio-2009}.

Convolutional neural networks (CNN) are a deep learning subfield that proved to be very effective for image classification, detection and segmentation. The first successful CNN was presented in \citep{LeCun:98}, and it was designed for hand-written digit recognition. This early CNN implementation used a combination of convolution, pooling and non-linearity that has been the key feature of DL since now. The DL breakthrough took place with the publication of \citep{NIPS2012_4824}, where for the first time a CNN won the Imagenet\citep{imagenet_cvpr09} classification competition by a large margin. In that paper a set of innovative techniques where introduced including data augmentation, the use of rectified linear units (ReLUs) as activation function, the use of dropout for avoiding overfitting, overlapping max-pooling for avoiding the averaging effects of avg-pooling and the use of GPUs for speeding up the training time. Later on, different CNN improvements where also published. In \citep{vggnet}, it was the first time that small 3x3 convolution filters where used and combined as a sequence of convolutions. In \citep{he2016deep} the authors introduced residual networks. This networks used a combination of 3x3 convolutional layers with a by-pass of the input every two layers that was summed up to the output. Such bypasses improved the gradient propagation through the network allowing the design of deeper networks and improving the classification capabilities. 

DL models have been also successfully applied in many medical classification tasks. In \citep{esteva2017dermatologist} a DL classifier was designed achieving dermatologist-level accuracy for skin cancer detection. In \citep{wentao2018deeplung} a 3D CNN for automated pulmonary nodule detection and classification was designed. In \citep{wang2018classification} a CNN Alzheimer's disease classifier with high performance was also described. In \citep{doi:10.1001/jama.2016.17216} the authors presented a diabetic retinopathy classifier with better performance than human experts in the detection of the most severe cases of the disease.

Ordinal regression is a term used for multi-class classification for the cases where a underlying property can be used for prestablishing an ordering of the classes. Several quality measures exist in the literature of machine learning and statistics for ordinal regression \citep{mehdiyev2016evaluating}. Kappa is a well-known statistic coefficient defined by Cohen \citep{cohen1960coefficient} to measure inter-rater agreement in such cases. Weighted Kappa \citep{cohen1968weighted} is another index used for measuring the goodness of a ordinal regression classification. In this last coefficient, disagreements are penalized proportionally to a power of the distance between classes. The penalization most commonly used is the quadratic. In such cases, the index is commonly referred as Quadratic Weighted Kappa (referred as QWK or $\kappa$ indistinctly in this paper).

In medical diagnosis tasks is important not only the accuracy of predictions but also the reasons behind decisions. Self-explainable models enable the physicians to contrast the information reported with their own knowledge, increasing the probability of a good diagnostic, which may have a significant influence in patient's treatment.  

Different attempts have been done for interpreting the results reported by neural networks. In \citep{zeiler2014visualizing} a network propagation technique is used for input-space feature visualization. After this work, \citep{bach2015pixel} used a pixel-wise decomposition for classification decision. This decomposition could be done in two ways: considering the network as a global function, disregarding its topology (functional approach) or using the natural properties of the inherent function topology for applying a message passing technique, propagating back into the pixel space the last layer output values. After this work, in \citep{montavon2017explaining} they used a so named Deep Taylor decomposition technique to replace the inherently intractable standard Taylor decomposition using a multitude of simpler analytically tractable Taylor decompositions.

Independent Component Analysis (ICA)\citep{hyvarinen2000independent} is a statistical method for the separation of a multi-dimensional random signal into a linear combination of components that are statistically as independent from each other as possible. The theoretical foundation of ICA is based on the Central Limit Theorem, which establishes that the distribution of the sum (average or linear combination) of N independent random variables approaches a gaussian as $N \rightarrow \infty$. When ICA method is applied, it is assumed that such separation exist, ie. that is possible to express the signal as a linear combination of independent components (IC). Perfect independence between random variables is achieved when mutual information between them is zero. Mutual information can be expressed as the Kullback-Leibler divergence between the joint distribution and the product of the distributions of each variable. Mutual information can be decomposed, under linear transforms, as the sum of two terms: one term expressing the decorrelation of the components and one expressing their non-Gaussianity \citep{cardoso2003dependence}. ICA uses optimization to calculate its components. Two types of optimization objectives can be used: minimize the mutual information or maximize the non-gaussianess of each component. 

In this paper we study a technique that allows the identification, separation and visualization in the input and hidden space of the IC responsible of a particular DR classification decision taken by a DL classifier given a certain eye fundus image. This is done by calculating the minimum number of IC that are able to encode the maximum information about the particular classification. Identifying such components, we reduce the redundancy of feature space and identify the components that share the minimum mutual information between each other. In that way, under the supposition that the feature extraction phase of the deep learning model has been able to disentangle completely the feature information, we are able to separate the independent elements causing the disease.

We use the pixel-wise visualization method to identify in pixel space such independent causes. We use the assessment of experts clinicians for interpreting such IC.

The paper is structured as follows: in Section \ref{sec:related} the current work on DL applied to DR is briefly introduced, then, the main works on interpretation of DL are presented. Section \ref{sec:methods} we present the methods, Section \ref{sec:results} presents the results showing samples of the kind of visual interpretations given by the model and finally Section \ref{sec:conclusions} present the final conclusions of our work.

\section{Related Work}\label{sec:related}

Many DL based DR classifiers have been published in the last years \citep{pratt2016convolutional}, \citep{DELATORRE2017}, \citep{doi:10.1001/jama.2016.17216}. 

In \citep{doi:10.1001/jama.2016.17216} a binary DL classifier was published for the detection of the most severe cases of DR (grouping classes 0 and 1 of DR on one side, and classes 2, 3 and 4 on another). This model was trained using an extended version of the EyePACS dataset mentioned before with a total of 128,175 images and improving the proper tagging of the images using a set of 3 to 7 experts chosen from a panel of 54 US expert Ophtalmologists. This model surpassed the human expert capabilities, reaching at the final operating point approximately  97\% sensitivity and 93.5\% specificity in the test sets of about 10,000 images for detecting the worse cases of DR. The strength of this model was its ability to predict the more severe cases with a sensitivity and specificity greater than human experts. 

In our previous work \citep{DELATORRE2017} a DL classifier was published for the prediction of five different disease grades. This model was trained using the public available EyePACS dataset. The training set had 35,126 images and the test set 53,576. The quadratic weighted kappa (QWK) evaluation metric \citep{cohen1968weighted} over the test set using a unique DL model without ensembling was close to the reported by human experts. 

The drawback of all these implementations, as of many other DL based models, is its lack of interpretability. In last years different approximations have been proposed for converting the initial DL black box classifiers into interpretable ones. Between the more successful interpretation models existing today we have the sensitivity maps \citep{DBLP:journals/corr/SimonyanVZ13}, layer-wise relevance propagation \citep{bach2015pixel} and Taylor type decomposition models \citep{montavon2017explaining}. 

All these methods use different strategies to backpropagate the classification scores to the input space and distribute the value of the final classification between input pixels. With this score distribution is possible to identify the most relevant pixels for a particular classification. 

Layer-wise relevance propagation models allow not only to report the predicted class but also to evaluate the importance of every input pixel of the image in the final classification decision. In such a way, it is possible to determine which pixels are more important in the final decision and facilitate the human experts the construction of rationale explanations based on the interpretation of such maps. 

\section{Methods}\label{sec:methods}

DL models are organized in layers, being the inputs of each one a combination of the outputs of previous ones. We design the output layer to be a linear combination of last layer feature space components. In this way we are forcing the model to disentangle the important features that, combined in a linear way, allow the achievement of a maximum possible classification score. These components (or other obtained as a linear combination of them, like with ICA analysis), are easy to analyze due to the linear nature of its relationship with the classification scores.

\subsection{ICA based interpretation procedure}

In this paper we go a step forward in the interpretation of score maps. Instead of generating directly the pixel maps associated with a particular class, we try to identify and separate independent elements associated with the disease. Our new contribution comes from the identification, separation and visualization of the IC that explain a particular classification decision. We hypothesize that in order to achieve high performance classification scores, the network has to encode the information required to make the classification. Human experts base its decisions in the number and types of lesions present in the image. That's why in some way, such information has to be present in a disentangled form in last layer feature space, previous to the output layer.  Instead of directly visualizing the more important pixels under a classification decision, we split the information of such last feature layer into independent features using a Independent Component Analysis (ICA). A posteriori we use a pixel-wise relevance propagation derived method to visualize such independent component in input space. In this way, we can, not only generate importance pixel maps, but also differentiate between the underlying independent causes of the disease.

We study the last layer feature space, previous to the output layer linear combination in order to identify its properties and try to isolate the independent elements that are causing a particular classification. For this purpose we use a principal component analysis (PCA)  \citep{pearson1901principal} to appraise the redundancy of this space and a ICA \citep{hyvarinen2000independent} using different number of components to identify the minimum of them required to achieve a classification score close enough to the achieved without such a dimensional reduction.  ICA allows to find a linear representation of non-Gaussian data so that the components are statistically independent, or as independent as possible. Such a representation seems to capture the essential structure of the data in many applications, including feature extraction \citep{hyvarinen2000independent}. When the data is not Gaussian, there are higher order statistics beyond variance that are not being taken into account by PCA. While PCA captures only uncorrelated components, these uncorrelated components are not necessarily independent for general distributions. ICA minimizes the mutual information (or relative Kullback-Leibler divergence) of non-Gaussian data because two distributions with zero mutual information are statistically independent \citep{comon1992independent}. 

\begin{figure}[h!]
	\centering
	\begin{subfigure}[b]{\textwidth}
		\centering
		\includegraphics[width=0.7\textwidth]{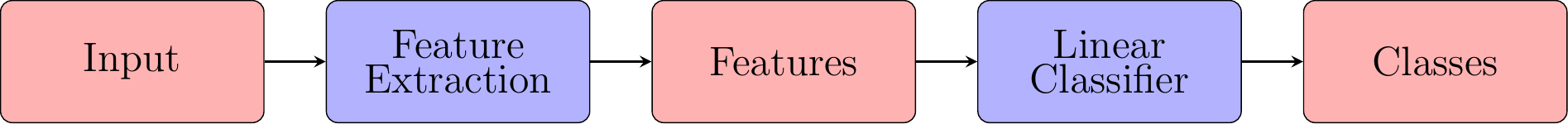}
		\caption{Initial classification network}	
	\end{subfigure}
	\hfill   
	\begin{subfigure}[b]{\textwidth}
		\centering
		\includegraphics[width=\textwidth]{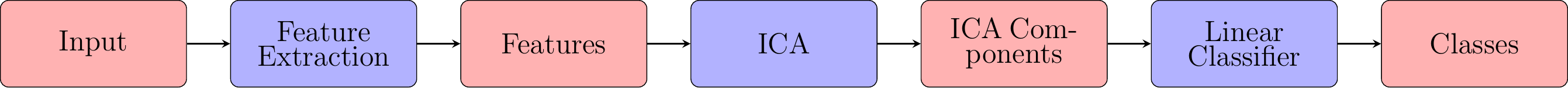}
		\caption{Modified for visualization of independent underlying classification causes}
		\label{fig:models_ica}
	\end{subfigure}
	\caption{Model changes done for improving explainability}  
	\label{fig:models} 
\end{figure}

For finding the optimal number of IC, we apply ICA to the last layer feature space training set vectors. Using different number of IC and comparing the classification performance of the original model with the obtained using a linear combination of the reduced number of calculated components, it is possible to find the optimal number of components ($n$) that does not significantly reduce the classification performance of the original model. We modify the original model adding a new layer after the last layer feature space to calculate online the components of every analyzed image. This layer is a linear transformation and acts as a dimensionality reduction layer (see fig. \ref{fig:models}). The final classification is achieved linearly combining the low dimensional IC layer. 

After identifying the optimal $n$, we use the receptive field and pixel-wise explanation model \citep{de2017deep} to visualize the independent scores in the input space. In this way we are visualizing not only a score map explaining a classification but also differentiating and visualizing the mathematically IC responsible of a particular classification. 

\subsection{Mathematical formalization}

Let $F_{train}$ be the set of all feature vectors of training set:

\begin{equation}
	F_{train} = \{\boldsymbol{f^{(i)}} : i = 1 .. T\}, \quad \boldsymbol{f^{(i)}} = (f^{(i)}_1, f^{(i)}_2, ... f^{(i)}_m)
\end{equation}

being $T$ the number of elements of the training set, and $m$ the dimension of feature space vector.

Let $S_{train}$ the set of IC calculated from $F_{train}$:

\begin{equation}
	S_{train} = \{\boldsymbol{s^{(i)}} : i = 1 .. T\}, \quad \boldsymbol{s^{(i)}} = (s^{(i)}_1, s^{(i)}_2, ... s^{(i)}_n)
\end{equation}

being $n$ the number of IC, $n < m$.

Every $\boldsymbol{s^{(i)}}$ can be expressed as a linear combination of $\boldsymbol{f^{(i)}}$:
\begin{equation}
\boldsymbol{s^{(i)}} = \boldsymbol{W} \boldsymbol{f^{(i)}}
\end{equation}

Where $\boldsymbol{W}$ is calculated using a optimization method, minimizing the mutual information between $n$ IC ($\min_{S} I(S)$) \citep{hyvarinen1999fast}.

The ordinal regression problem solved using $F_{train}$ can be expressed as:

\begin{equation}
\max_{\boldsymbol{A}} \big[ \kappa_{val} (C_{train}) \big]
\end{equation}

being:

\begin{equation}
C_{train} = \{ \boldsymbol{A} \boldsymbol{f^{(i)}}, \forall \boldsymbol{f^{(i)}} \in F_{train} \} 
\end{equation}

being $\boldsymbol{c^{(i)}} = \boldsymbol{A} \boldsymbol{f^{(i)}}$ the predicted class vector and $\kappa_{val}$ the evaluation function used in the training process of the neural network calculated for the validation set.

We want to solve the same problem using the reduced ICA components (feature space compressed version):

\begin{equation}
C'_{train} = \{ \boldsymbol{B} \boldsymbol{s^{(i)}}, \forall \boldsymbol{s^{(i)}} \in S_{train} \} 
\end{equation}

being $\boldsymbol{c'^{(i)}} = \boldsymbol{B} \boldsymbol{s^{(i)}}$ the predicted class vector using $n$ IC.

The optimal number of IC to select is the one that minimizes the difference in performance between both models:

\begin{equation}
\min_{n} \big[ \kappa_{val} (C'_{train}) - \kappa_{val} (C_{train}) \big] 
\end{equation}

\section{Results}\label{sec:results}

\subsection{Data}

In this study we use the EyePACS dataset of the Diabetic Retinopathy Detection competition hosted on the internet Kaggle Platform.  For every patient right and left eye images are reported. All images are classified by ophthalmologists according to the standard severity scale presented before in \citep{diaclass}. The images are taken in variable conditions: by different cameras, illumination conditions and resolutions. 

The training set contains a total of $75,650$ images; $55,796$ of class 0, $5,259$ of class 1, $11,192$ of class 2, $1,805$ of class 3 and $1,598$ of class 4. The validation set used for hyper-parameter optimization has $3,000$ images; $2,150$ of class 0, $209$ of class 1, $490$ of class 2, $61$ of class 3 and $90$ of class 4. The test set, contains a total of $10,000$ images; $7,363$ of class 0, $731$ of class 1, $1,461$ of class 2, $220$ of class 3 and $225$ of class 4. 

\subsection{Baseline Model}

Our baseline model uses a 3x640x640 input image obtained from a minimal preprocessing step where only the external background borders are trimmed and later resized to the required input size. It is a CNN of 391,325 parameters, divided in 17 layers. Layers are divided in two groups: the feature extractor and the classifier. The feature extraction has 7 blocks of 2 layers. Every layer is a stack of a 3x3 convolution with stride 1x1 and padding 1x1 followed by a batch normalization and a ReLU activation function. Between every block a 2x2 max-pooling operation of stride 2x2 is applied.  Afterwards, the classification phase takes place using a 2x2 convolution. A 4x4 average-pooling reduces the dimensionality to get a final 64 feature vector that are linearly combined to obtain the output scores of every class. A soft-max function allows the conversion of scores to probabilities to feed the values to the proper cost function during the optimization process. The feature extractor has 16 filters in the first block, 32 in the second and 64 in all the other.

\subsection{Model Modifications}

For all the training set we calculate the last layer feature space, obtaining a 64-dimensional vector as a representation of each image. We observe that this vector is highly redundant. After applying a PCA, with only 10 components it is possible to explain 99\% of the variance.

\begin{figure}[h]
	\centering	
	\includegraphics[width=0.65\textwidth]{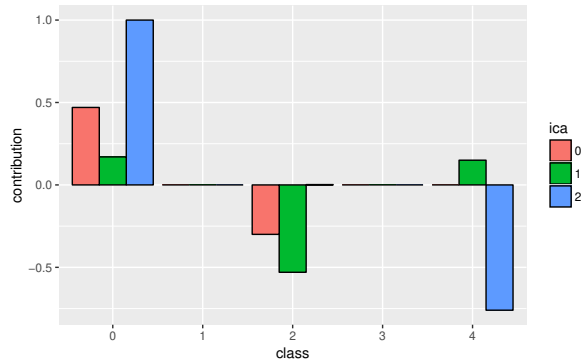}
	\caption{Contribution of each ICA component in the classification final score}
	\label{fig:ica_contribution}
\end{figure}

Using the 64-dimensional feature-space vector ($m=64$) of all the training set, we calculate a set of ICAs using different $n$ values. With each one, we train a linear classifier to calculate the evaluation metric obtained over a validation set (fig. \ref{fig:models_ica}). We choose the minimal $n$ that allows achieving maximum performance. The optimal $n$ for this problem is $3$, achieving a $QWK_{val} = 0.790$ not far from the achieved by the original model without dimensionality reduction ($QWK_{val} = 0.800$).

Fig. \ref{fig:ica_contribution} shows the contribution of each component to the score of each class. We can see that ICA values are differentiating between extreme classes 0, 2 and 4. As we are training the network using as a loss function $QWK$, the optimization takes place as an ordinal regression. We expect the network to find the underlying causes present in the image that produce the predefined sorting of the classes, ie. the types of lesions and the number of them. Class 0 score contributions come from $ICA_0 > 0$, $ICA_1 > 0$ and $ICA_2 > 0$; being the class markers of the presence of disease $ICA_0 < 0$, $ICA_1 < 0$ and $ICA_2 < 0$. Analyzing the pixels with higher negative signals in the three components will give us the points that are contributing the most to the signaling of a possible presence of the disease. Backpropagating the scores of each one of this negative components will give a richer explanation with distinction between three possible independent causes of the final diagnostic given by the model.

\begin{figure}[h]
	\centering
	\begin{subfigure}[b]{0.49\textwidth}
		\centering
		\includegraphics[width=\textwidth]{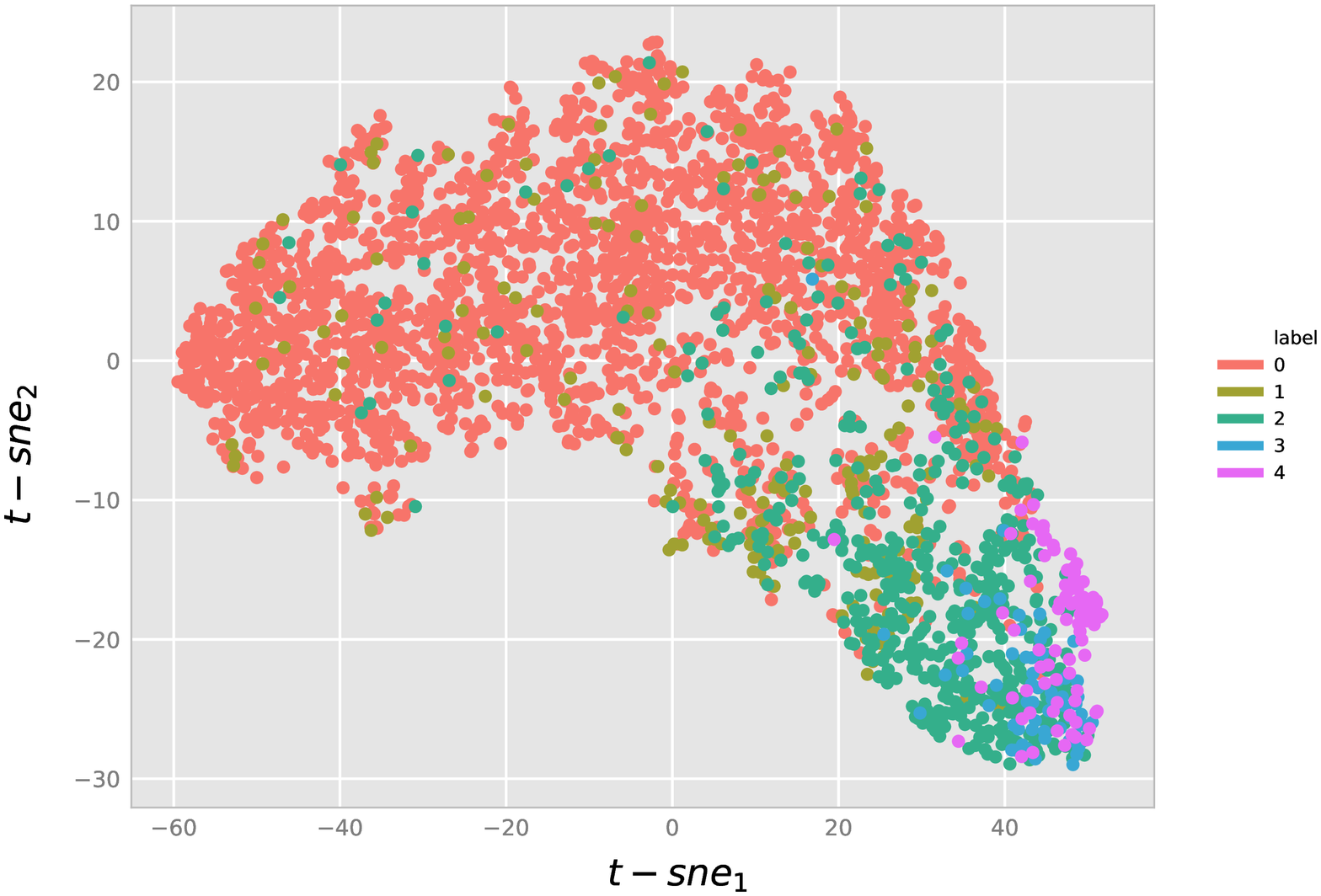}
		\caption{Original feature space (64 components)}	
	\end{subfigure}
	\begin{subfigure}[b]{0.49\textwidth}
		\centering
		\includegraphics[width=\textwidth]{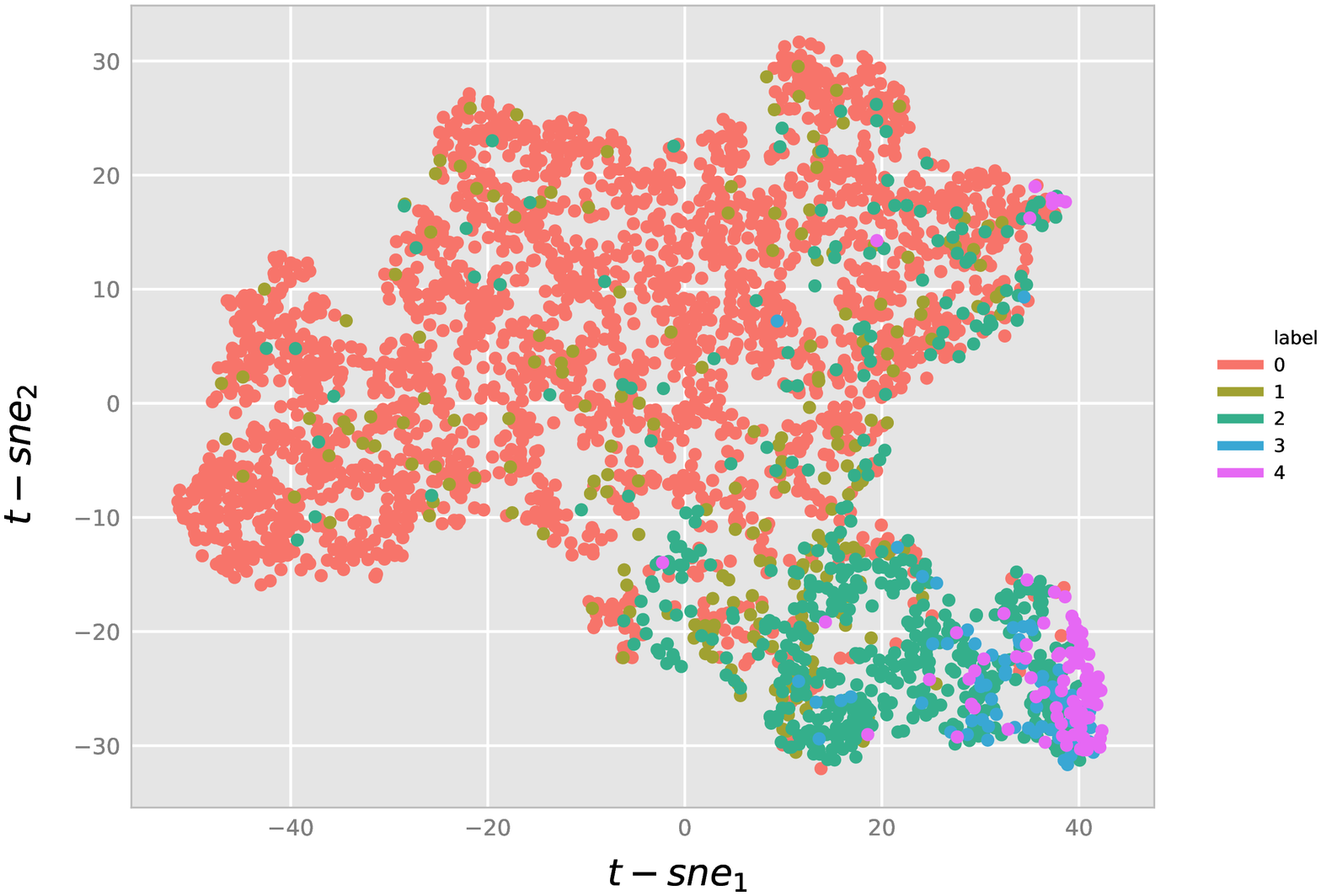}
		\caption{ICA space (3 components)}
	\end{subfigure}
	
	\caption{Comparison between the 2D t-SNE visualization of validation set using the original feature space and the final 3-dimensional ICA space}  
	\label{fig:tsne} 
\end{figure}

We use a two-dimensional t-SNE visualization \citep{maaten2008visualizing} for qualitative evaluation of the difference between the quality of the separation using the 64 feature vector and of the reduced version with only 3 ICA components. In fig. \ref{fig:tsne} we show the 2D t-SNE visualization for the original feature space and for the ICA reduced space. We can see how for both spaces class 0, 2 and 4 classes are clearly separated. Class 0 and 1 are not properly separated and in the case of 3 and 4, although the separation is not perfect, it is possible to distinguish a different location of both classes for both spaces. As the quantitative index QWK also show, there is not a qualitative separation capability difference between the t-SNE map of the original features and the map of the ICA 3 dimensional compressed feature map. For such $1.25\%$ difference in the classification performance index ($\kappa_{orig} = 0.800$ vs $\kappa_{ICA} = 0.790$) we can conclude that the independent component analysis has been able to find a adequate compressed expression of the information encoded in the network. 

\subsection{Score components contribution for a test sample}

\begin{figure}[h!]
	\centering
	\begin{subfigure}[b]{0.32\textwidth}
		\centering
		\includegraphics[width=\textwidth]{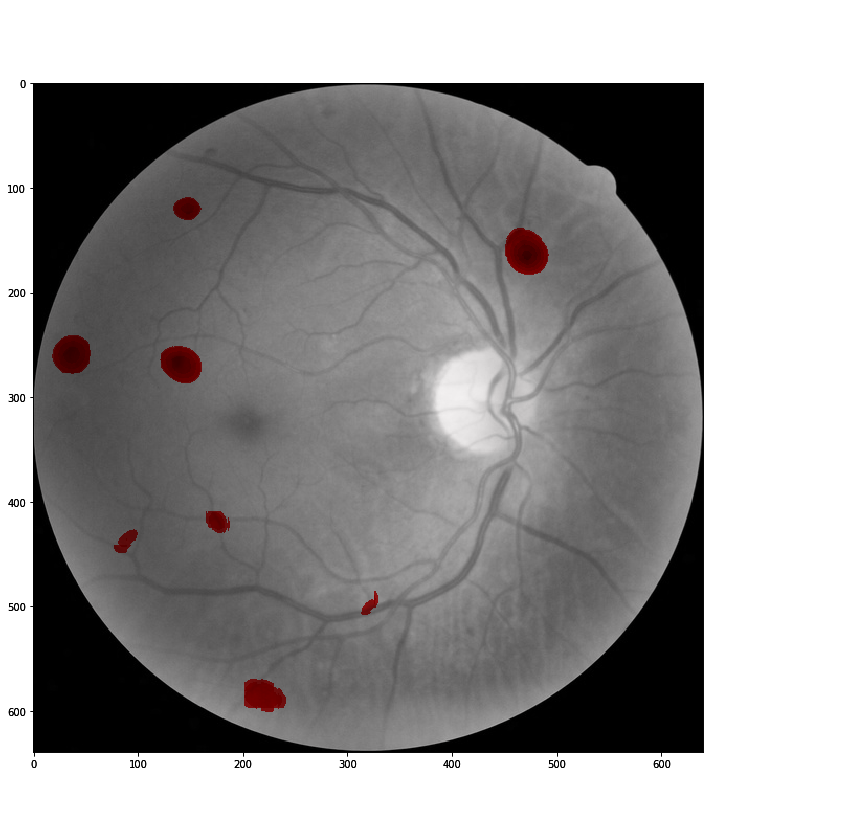}
		\caption{$ICA_0 < - 2 \sigma$}	
	\end{subfigure}
	\begin{subfigure}[b]{0.32\textwidth}
		\centering
		\includegraphics[width=\textwidth]{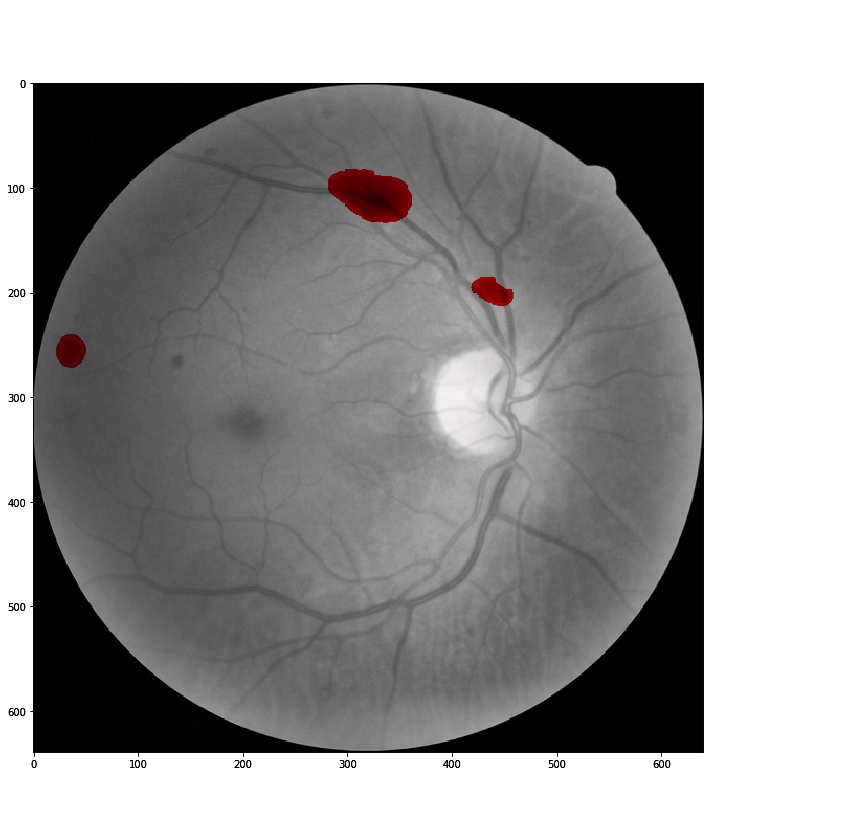}
		\caption{$ICA_1 < - 2 \sigma$}	
	\end{subfigure}
	\begin{subfigure}[b]{0.32\textwidth}
		\centering
		\includegraphics[width=\textwidth]{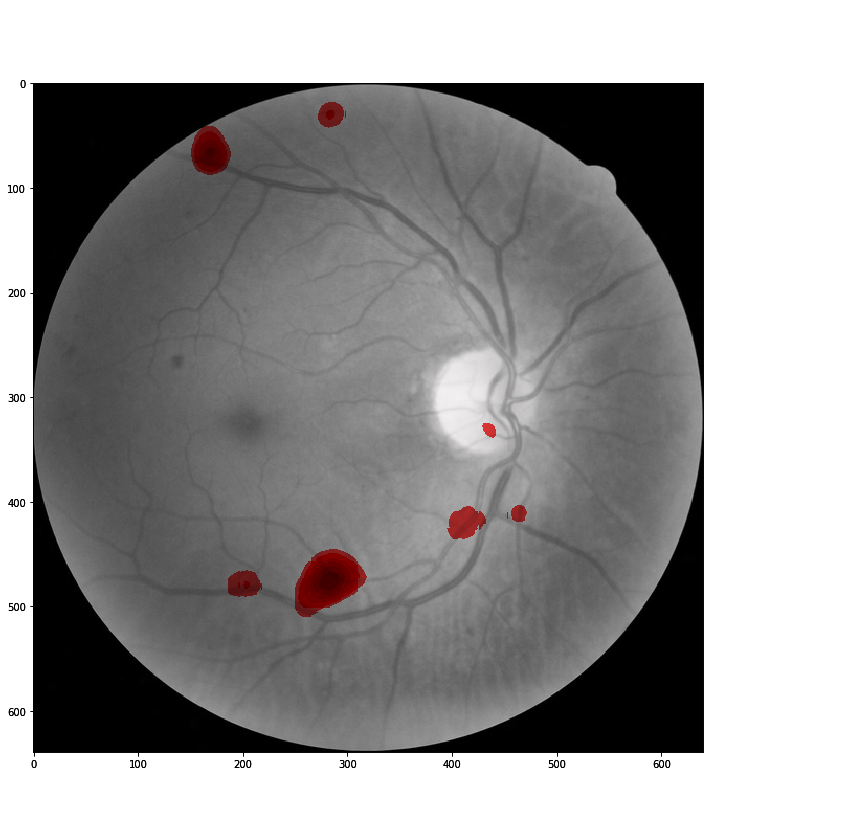}
		\caption{$ICA_2 < - 2 \sigma$}	
	\end{subfigure}
	\hfill 
	\caption{ICA score maps generated for a class 2 image (predicted also as class 2) using a receptive field of 61x61}  
	\label{fig:ica_components_class2} 
\end{figure}

\begin{figure}[h!]
	\centering
	\begin{subfigure}[b]{0.32\textwidth}
		\centering
		\includegraphics[width=\textwidth]{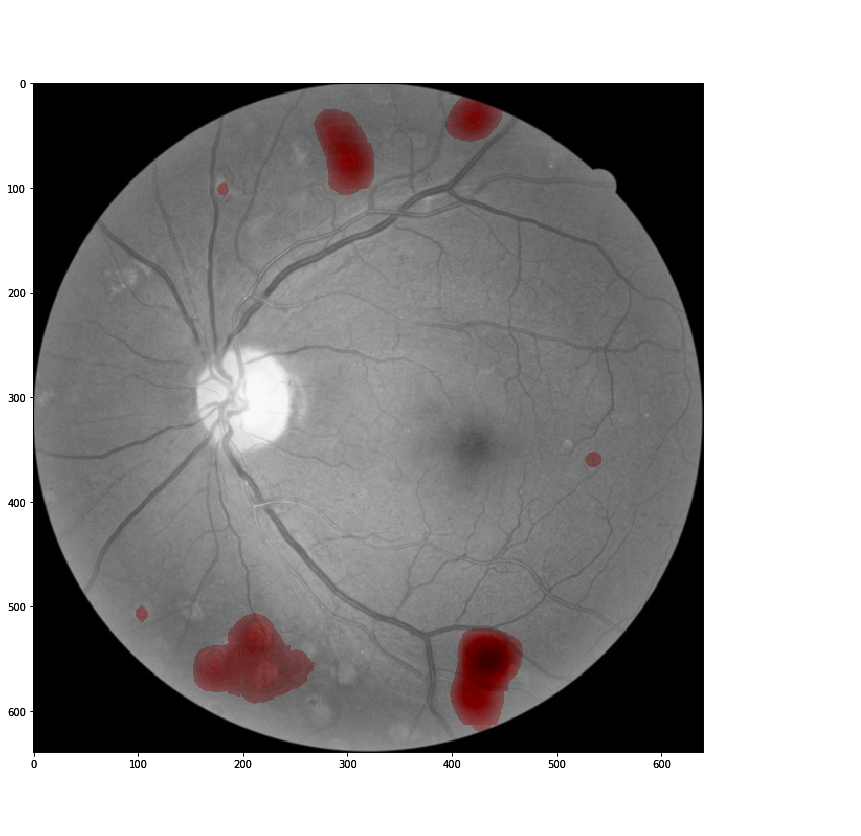}
		\caption{$ICA_0 < - 2 \sigma$}	
	\end{subfigure}
	\begin{subfigure}[b]{0.32\textwidth}
		\centering
		\includegraphics[width=\textwidth]{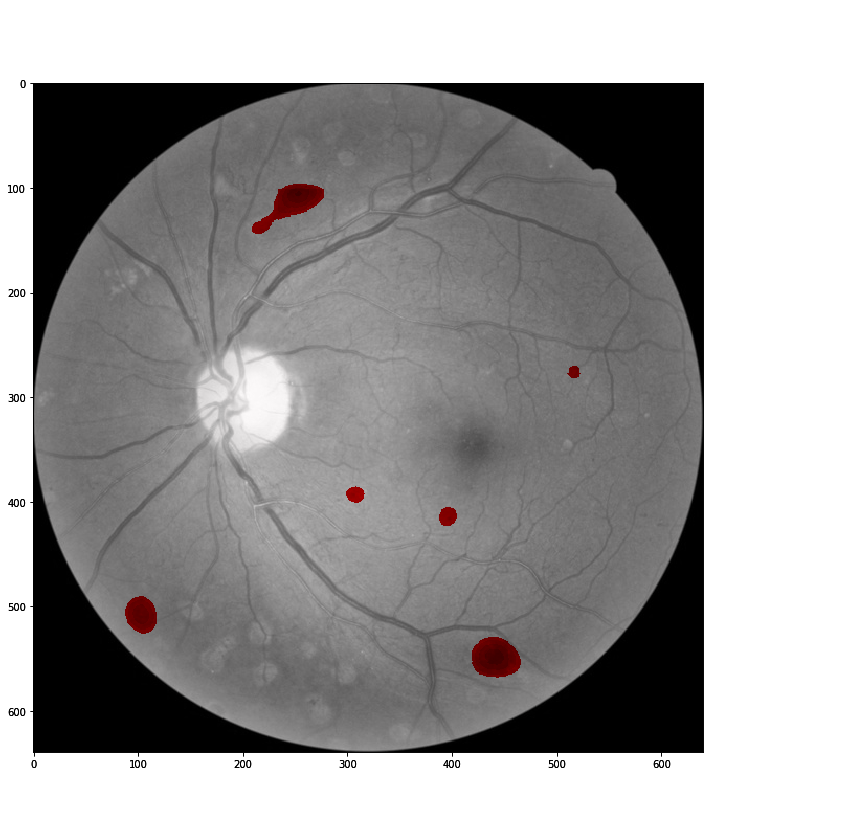}
		\caption{$ICA_1 < - 2 \sigma$}	
	\end{subfigure}
	\begin{subfigure}[b]{0.32\textwidth}
		\centering
		\includegraphics[width=\textwidth]{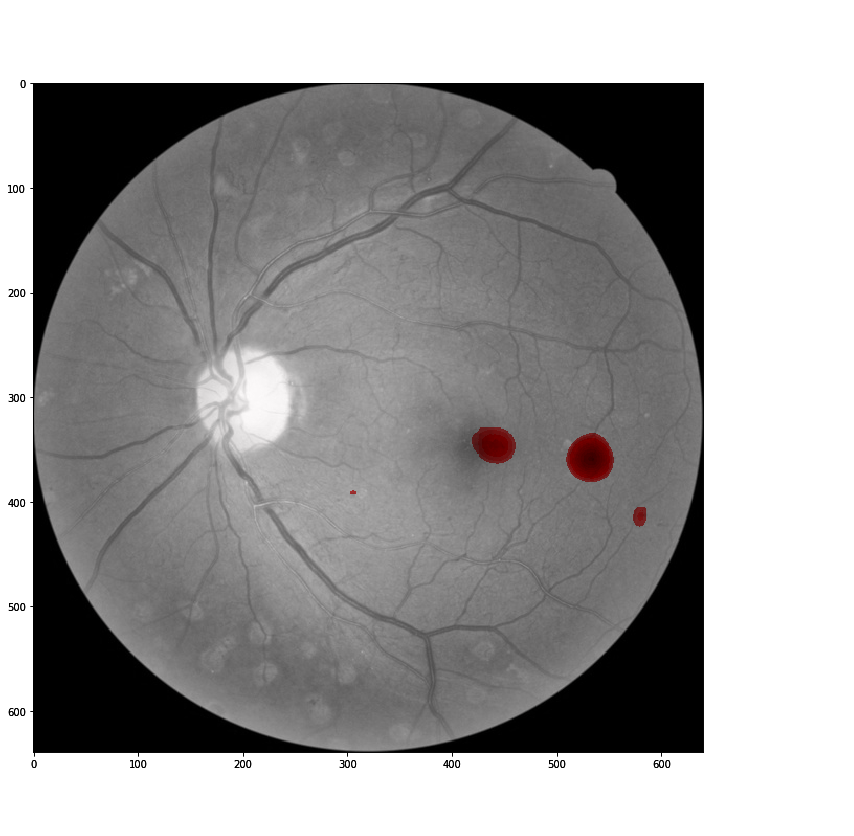}
		\caption{$ICA_2 < - 2 \sigma$}	
	\end{subfigure}
	\hfill 
	\caption{ICA score maps generated for a class 4 image (predicted also as class 4) using a receptive field of 61x61}  
	\label{fig:ica_components_class4} 
\end{figure}

We use a pixel-wise relevance propagation derived method for visualizing each ICA component independently. In this way, it is possible to visualize the mathematically independent contributions, enhancing the localization of different types of primary elements causing the disease. Figure \ref{fig:ica_components_class2} shows the intermediate score maps generated using a receptive field of 61x61. Figure \ref{fig:ica_components_class4} shows the same intermediate scores for a class 4 image. These hidden layer maps are useful when a general map of the lesion locations is enough.

Input-space pixel scores are the most suitable when pixel detail is required for detecting the individual lesions causing the disease.

\begin{figure}[h!]
	\centering
	\begin{subfigure}[b]{\textwidth}
		\centering
		\includegraphics[width=\textwidth]{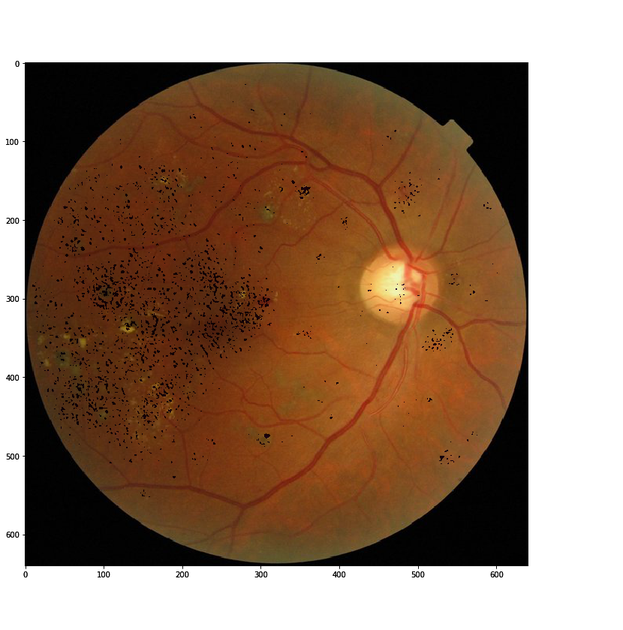}
	\end{subfigure}
	\caption{Visualization of $ICA_2$ score map for a class 3 image ($-ICA_2 < 2 \sigma_2$). The classification model scores are $C_0 = -661.0$, $C_1 = -294.1$,  $C_2 = -10.3$,   $C_3 = 70.3$,   $C_4 = 26.3$; $ICA_0 = 0.0174$, $ICA_1 = -0.0181$, $ICA_2 = -0.0237$}  
	\label{fig:ica_components_c3} 
\end{figure}

Figure \ref{fig:ica_components_c3} show the input space final scores for $ICA_2$. We show the points having contributions higher than a prestablished limit, in this case with values higher than two standard deviations. 

$ICA_2$ score maps have an almost perfect match with real lesions having a very low rate of false positives and false negatives. $ICA_0$ and $ICA_1$ include statistical regularities not directly related with relevant lesions. From such interpretation we conclude that ICA acts as a filter separating lesion information present in images from blink artifacts, noise and other statistical regularities present in images.

\section{Conclusions}\label{sec:conclusions}

In this paper we studied some of the feature space properties of a already trained diabetic retinopathy deep learning classifier of 17 layers. We saw how, even being a small network, the redundancy of the feature space is very high. We hypothesized that if the network is able to achieve human-level classifications, in some way, it has been able to identify the important statistical regularities present in the image that are important for the classification. As last layer is a linear classifier, such properties has to be disentagled, ie. expressed in a linear way in such last layer. We applied a independent component analysis over such last layer with the method explained in the paper in order to find the optimal number of components that maximize the classification capabilities of such compressed version of the features. We experimentally proved that reducing to only 3 components is possible to achieve almost $99\%$ of the evaluation metric. Such value experimentally prove that the ICA compression has been able to extract all the important elements required for the classification. With such elements, we applied a visualization technique for identifying the elements in the image of each one of the components, using a pixel-wise derived visualization technique. We are able to generate three maps for every image each one identifying independent statistical regularities important for the classification.

Our method allows not only the classification of retinographies but also the identification and localization in the image of the mathematically independent signs of the disease. The presented ICA score model is of general applicability and can be easily adapted for the usage in other image classification tasks thus, as future work we plan to test it in other types medical images (eg. cancer detection in mammograms).

\section*{Acknowledgements}

This work is supported by the URV grants 2017PFR-URV-B2-60, and the Spanish research project PI15/01150 (Instituto de Salud Carlos III) and Fondos FEDER. The authors would like to thank to Kaggle and EyePACS for providing the data used in this paper.

\section*{References}

\end{document}